\documentclass[letterpaper, 10 pt, conference]{ieeeconf}  

\IEEEoverridecommandlockouts   

\usepackage{graphics} 
\usepackage{epsfig} 
\usepackage{mathptmx} 
\usepackage{times} 
\usepackage{amsmath} 
\usepackage{amssymb}  
\usepackage{multirow}

\newcommand{\p}{\Delta \mathbf{p}}
\newcommand{\R}{\Delta \mathbf{R}}
\newcommand{\bp}{\overline{\Delta \mathbf{p}}}
\newcommand{\bR}{\overline{\Delta \mathbf{R}}}

\usepackage{color}

\usepackage{bm}

\usepackage[normalem]{ulem}

\title{\LARGE \bf
Invariant Extended Kalman Filtering for Human Motion Estimation with Imperfect Sensor Placement
}

\author{Zenan Zhu$^{1,*}$, Seyed Mostafa Rezayat Sorkhabadi$^{2,*}$, Yan Gu$^{1,\dagger}$ Wenlong Zhang$^{2}$
\thanks{This work was supported by the National Science Foundation under Grants IIS-1756031, IIS-1955979, CMMI-1944833, and CMMI-2046562.}
\thanks{$^{1}$Z. Zhu and Y. Gu are with the College of Engineering,
        University of Massachusetts Lowell, Lowell, MA 01854, USA
        {\tt\small zenan\_zhu@student.uml.edu, yan\_gu@uml.edu.}}%
\thanks{$^{2}$M. Rezayat and W. Zhang are with The Polytechnic School, Ira A. Fulton Schools of Engineering, Arizona State University
        Mesa, AZ 85212, USA
        {\tt\small \{sm.rs, wenlong.zhang\}@asu.edu}.}%
\thanks{$*$ These two authors have equal contributions.}
\thanks{$\dagger$ Corresponding author: Y. Gu.}
}

\begin{document}

\maketitle

\thispagestyle{empty}
\pagestyle{empty}

\begin{abstract}
This paper introduces a new invariant extended Kalman filter design that produces real-time state estimates and rapid error convergence for the estimation of the human body movement even in the presence of sensor misalignment and initial state estimation errors.
The filter fuses the data returned by an inertial measurement unit (IMU) attached to the body (e.g., pelvis or chest) and a virtual measurement of zero stance-foot velocity (i.e., leg odometry).
The key novelty of the proposed filter lies in that its process model meets the group affine property while the filter explicitly addresses the IMU placement error by formulating its stochastic process model as Brownian motions and incorporating the error in the leg odometry.
Although the measurement model is imperfect (i.e., it does not possess an invariant observation form) and thus its linearization relies on the state estimate, experimental results demonstrate fast convergence of the proposed filter (within 0.2 seconds) during squatting motions even under significant IMU placement inaccuracy and initial estimation errors. 

\end{abstract}

\section{INTRODUCTION}

Over the past decades, wearable robots have become increasingly applied in daily living assistance, neurorehabilitation, and power augmentation~\cite{deng2018advances}. For a wearable robot to function autonomously with different users in various tasks, it needs to collect sensor data to understand human states and intents in real time. For lower-extremity wearable robots, human data have been primarily gathered using wearable sensors, including optical encoders, inertial measurement units (IMU), surface electromyography (EMG) sensors, to name a few. However, a significant amount of useful information cannot be directly measured by sensors since it is impractical to implant sensors inside the human body. In this case, it is critical to define and estimate human states in various human-robot interactive tasks.   

For human locomotion, both continuous and discrete states have been introduced to quantify and differentiate various motion patterns and intents. Gait phases have been widely used to describe the human states during walking: a typical step involves stance and swing phases, each with multiple sub-phases defined~\cite{whittle2014gait}. Besides gait phases, wearable robots also need to estimate the human motion intents, such as sitting, walking, standing, and stair-climbing~\cite{zhang2020unsupervised}. Gait phases and motion intents are often finite, and they can be estimated from lower-extremity joint angles and ground reaction forces using finite-state machines~\cite{lawson2012control}, fuzzy logic~\cite{chinimilli2017human}, and learning-based classifiers~\cite{su2019cnn}. In contrast, continuous human states are often estimated by sensor fusion and regression models for improved accuracy. 
One variable of particular interest in continuous state estimation is a person's stance-foot position in the world,
which can be used to represent a locomotor's global position in an environment~\cite{ojeda2007personal}.
With an IMU attached to each toe, the dead reckoning method [12] obtains the toe velocity by integrating the accelerometer reading, and removes the accumulated velocity errors due to the integration by resetting the velocity to zero when the toe is static on the ground.
The dead reckoning method has been applied to achieve real-time human localization~\cite{van2016real}, and extended to further improve its accuracy through smoothing~\cite{ruiz2011accurate} or filtering~\cite{ruiz2011accurate}.

Besides the stance-foot location, the pose (position and orientation) and velocity of the body (e.g., pelvis or chest) are also of particular interest in gait analysis and wearable robot controller design, because they can be used to study postural balance and gait stability~\cite{deane2021pressure}.
Body pose and velocity have been estimated through the nonlinear forward kinematics between the stance foot, which is obtained through accurate initialization and contact detection, and the body frame~\cite{yuan20133}.
This method assumes the leg kinematics is precisely known, and thus has been extended based on Kalman filtering (KF) to explicitly address uncertainties such as sensor noises~\cite{yuan2014localization}.
Recently, extended Kalman filtering (EKF) has been applied to further address the inaccuracy of the nonlinear kinematics chain, in addition to sensor imperfections, for real-time movement estimation under small initial estimation errors~\cite{zhang2013rider,zhang2015whole}.
Yet, conventional EKF suffers the major weakness that its design relies on the linearization of process and measurement models at the state estimates instead of the true states.
Due to this weakness, the EKF cannot provably guarantee error convergence in the presence of large estimation errors.

Recently, invariant extended Kalman filtering (InEKF) has been introduced to ensure real-time, provable error convergence even in the presence of large initial estimation errors~\cite{barrau2016invariant}.
The InEKF exploits nonlinear state estimation errors that are invariant on the matrix Lie group, and ensures that the dynamics of the logarithmic error is exactly linear and independent of the state estimate if the process model meets the group affine condition and if the measurement model is in the invariant observation form.
The filtering method has been applied to solve the real-time state estimation problem for aircraft~\cite{barrau2016invariant}, legged robots~\cite{hartley2020contact,lin2021deep,gao2021invariant,teng2021legged}, and underwater vehicles~\cite{potokar2021invariant}. 

While the InEKF method~\cite{hartley2020contact} has achieved impressive estimation performance for robot locomotion, it has not been applied to solve some of the key challenges in the state estimation of continuous human movement state, such as the inaccurate kinematic parameters and imperfect sensor placement.
One common solution to imperfect sensor placement is manual sensor calibration~\cite{yuan20133}, which is often time-consuming and thus may not be suitable for real-world applications (e.g., daily movement monitoring) that could demand frequent re-calibration.
Motivated by the practical demand of addressing inaccurate sensor placement, this paper introduces an InEKF that produces real-time state estimates and rapid error convergence of the body's pose and movement even in the presence of significant sensor placement offsets and large initial state estimation errors.
The key novelty of the proposed filter lies in that its process model meets the group affine property while the filter explicitly addresses the IMU placement error by formulating its stochastic process model as Brownian motions and incorporating the error in the leg odometry.

The rest of the paper is structured as follows.
Section II introduces the problem formulation.
Section III presents the proposed InEKF design with explicit treatment of sensor placement errors.
Section IV reports the experimental setup and validation results.
Section V provides the concluding remarks and future research directions.

\section{PROBLEM FORMULATION}

In this section, we will introduce an InEKF to estimate the states of the body (e.g., pelvis or chest) and the IMU placement offset, by using an IMU placed on the body to form the process model and by exploiting the stance leg's forward kinematic velocity to build a measurement model.

\begin{figure}[t]
    \centering
    \includegraphics[width=1\linewidth]{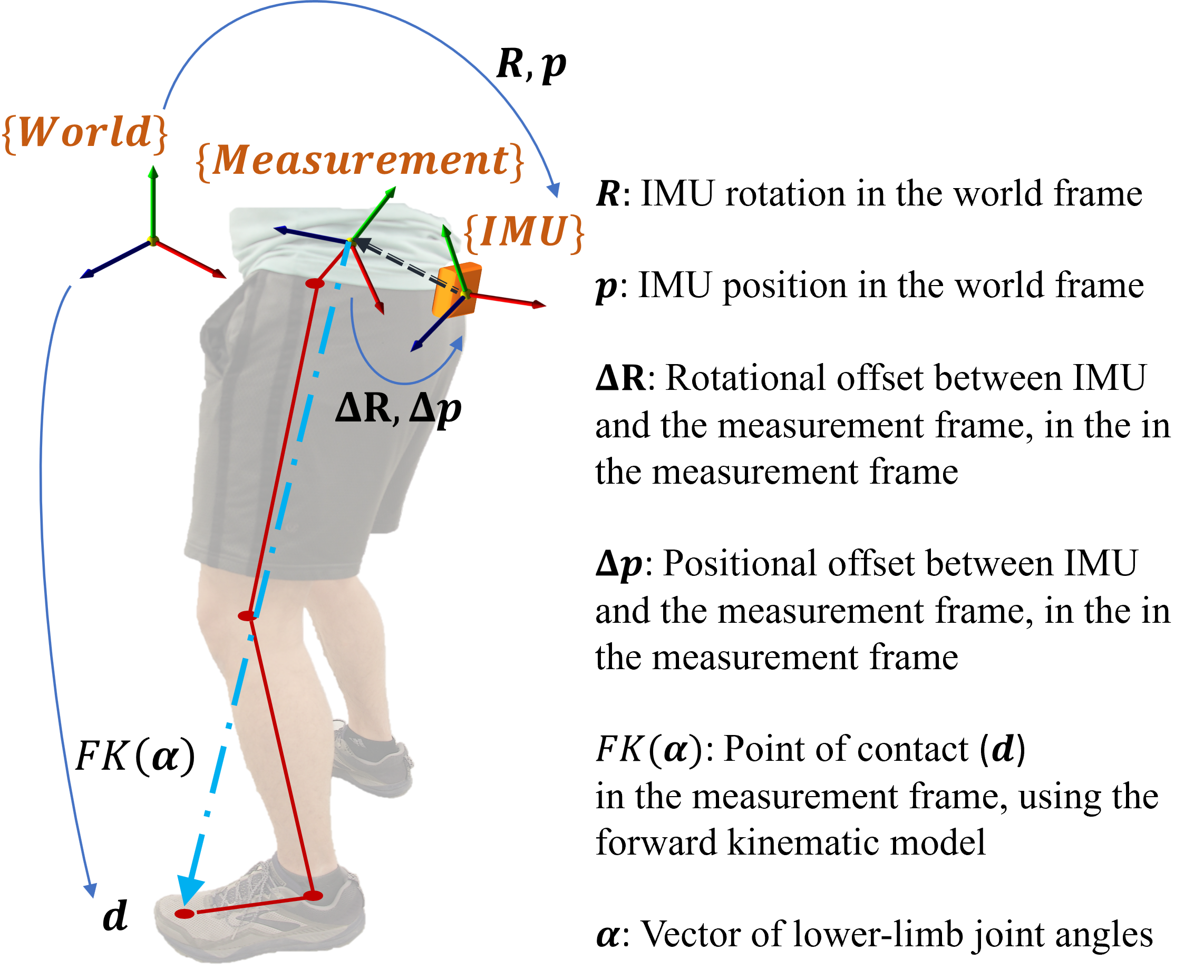}
    \caption{Measured and estimated variables in the proposed human body movement estimation.
    $\{World\}$ represents the world frame and $\{Measurement\}$ is the frame at which the measurements are provided. For the leg forward kinematic measurements, the measurement frame is defined at the center of pelvis.
    $\{IMU\}$ is the frame attached to the IMU. This figure illustrates the rotational and positional offset between the measurement and IMU frames, as well as the forward kinematics chain.
    }
    \label{fig: parameters}
\end{figure}

The human forward kinematic model provides the contact point position in the measurement frame (as shown in Fig.~\ref{fig: parameters}).
Thus there are some orientation and positional offsets between the measurement frame, which is considered in the human forward kinematic model, and where IMU is placed. 
The ``perfect" placement of the IMU would allow the exact alignment between the IMU and the measurement frames. 
Yet, such a placement is difficult to achieve in real-world applications without resorting to careful and often time-consuming manual calibration.
Here, to address the ``imperfect'' alignment between frames, we include the (orientation and positional) placement offset as part of the state estimation to 
make the connection between the process model (IMU) and the measurement more accurate.
This will lead to more accurate estimation of the body pose, and eliminate the need for accurate calibration between the IMU and the measurement frames.

\subsection{Process and Measurement Models} \label{Process}

Since the filtering objective is to estimate the body movement, we choose the state variables of the filter system to be the position $\mathbf{p}\in \mathbb{R}^3$, velocity $\mathbf{v}\in \mathbb{R}^3$, and orientation $\mathbf{R}\in SO(3)$ of the IMU, which is placed on the body, expressed in the world frame.
In addition, to explicitly treat the IMU placement offsets, the state variables also include the positional offset $\p \in \mathbb{R}^3$ and orientation offset $\R \in SO(3)$ of the IMU frame, represented in the measurement frame. 

\subsubsection{Process model}
The process model is based on the IMU motion characteristics. The IMU measures the linear acceleration $\mathbf{a} \in \mathbb{R}^3$  and angular velocity $\boldsymbol{\omega} \in \mathbb{R}^3$ in the IMU frame.
The sensor readings $\mathbf{\tilde{a}}$ and $\boldsymbol{\tilde{\omega}}$ are corrupted by zero-mean Gaussian white noise $\mathbf{w}_a$ and $\mathbf{w}_{\omega}$:
\begin{equation}
    \begin{gathered}
\mathbf{\tilde{a}}= \mathbf{a}+\mathbf{w}_a, \quad
\boldsymbol{\tilde{\omega}}= \boldsymbol{\omega}+\mathbf{w}_{\omega}. 
    \end{gathered} \label{IMU-input-nobias}
\end{equation}
Note that for simplicity, the biases in the raw data returned by the IMU are not considered here. Such biases could be treated by including them in the state and explicitly considering their dynamics~\cite{hartley2020contact}.

Considering these measurements as the input of the IMU motion dynamics, the process model becomes: 
\begin{equation}
    \begin{gathered} 
\frac{d}{dt}\mathbf{R}= \mathbf{R}(\tilde{\boldsymbol{\omega}}-\mathbf{w}_{\omega})_\times,
~
\frac{d}{dt}\mathbf{p}=\mathbf{v},
~
\frac{d}{dt}\mathbf{v}=\mathbf{R}(\tilde{\mathbf{a}}-\mathbf{w}_a)+\mathbf{g},  \label{IMU-dynamics-nobias}\\
    \end{gathered}
\end{equation}
where $(.)_\times$ denotes a skew-symmetric matrix and $\mathbf{g}$ is the gravitational acceleration vector. 

As the IMU placement offsets are typically constant, we model their dynamics as zero plus small zero-mean white Gaussian noise $\mathbf{w}_{\Delta p}$ and $\mathbf{w}_{\Delta R}$:
\begin{equation}
    \begin{gathered} 
\frac{d}{dt}{\p}=\mathbf{w}_{\Delta p}, \quad \frac{d}{dt}{\R}={\R}(\mathbf{w}_{\Delta R})_\times. \label{offset-dynamics}
    \end{gathered}
\end{equation}

\subsubsection{Leg kinematics measurement} \label{sec:Formulation_Leg_kinematics}
When either of the human feet is in a secured contact with the ground, we can estimate the position of one of many contact points (for example, toe) in the measurement frame, using the forward kinematics, which requires knowledge of the joint angles and the links depicted in Fig.~\ref{fig: parameters}. 
The forward kinematic model can be built using the Denavit-Hartenberg (DH) table approach \cite{denavit1955kinematic}. 
 
The contact point position in the measurement frame $\mathbf{d}^{M}\in \mathbb{R}^3$ is then acquired by the forward kinematic model as shown in Fig.~\ref{fig: parameters}: 
\begin{equation}
\mathbf{d}^{M}={FK}(\mathbf{\boldsymbol{\alpha}}). \label{leg_kinmetics}
\end{equation}

Later in our filter design (Sec.~\ref{sec: design}) we show how we use this model to get the contact-point velocity update equation that explicitly incorporates the IMU placement offsets and helps ensure some important properties for the filter.

\section{InEKF DESIGN with IMPERFECT SENSOR PLACEMENT}
\label{sec: design}
In this section we show the design of the proposed filter, incorporating the measurement frame positional and rotational offsets into the InEKF defined on a Lie group.
 
\subsection{State Representation and Propagation}
\label{sec:design_prop}
The first step to design an InEKF is to define the states on a matrix Lie group $G$~\cite{howe1983very, sola2018micro}, with its associated Lie algebra $\boldsymbol{\mathfrak{g}}$.
Here the variables we wish to estimate (introduced in Sec. \ref{Process}) are represented on a matrix Lie group: 
\begin{equation}
    \mathbf{X}=\begin{bmatrix}
\mathbf{R} & \mathbf{v} & \mathbf{p} & \mathbf{0}_{3,3} & \mathbf{0}_{3,1}\\
\mathbf{0}_{1,3} & 1 & 0 & \mathbf{0}_{1,3} & \mathbf{0}_{3,1}\\
\mathbf{0}_{1,3} & 0 & 1 & \mathbf{0}_{1,3} & \mathbf{0}_{3,1}\\
\mathbf{0}_{3,1} & \mathbf{0}_{3,1} & \mathbf{0}_{3,1} & \R & \p\\
\mathbf{0}_{1,3} & 0 & 0 & \mathbf{0}_{1,3} & 1
\end{bmatrix}
\in G,
\label{eqn: X}
\end{equation}
where the matrix Lie group $G$ is an extension of the special Euclidean group $SE(3)$, and $\mathbf{0}_{n,m}$ represents an $n\times m$ matrix with all elements being zero.
The proof that $G$ is a matrix Lie group is omitted due to space limit.

The core idea of the InEKF is the invariant error definition. The right-invariant error $\boldsymbol{\eta}$ between the true and estimation values is defined as~\cite{barrau2016invariant}: 
\begin{equation}
\boldsymbol{\eta}=\Bar{\mathbf{X}}\mathbf{X} ^{-1}\in G,
\end{equation}
where $\overline{(\cdot)}$ denotes the estimated value of the variable $(\cdot)$.

The tangent space $\mathfrak{g}$ (defined at the identity element $\mathbf{I} \in G$) is a vector space that can also be represented by vectors in the Cartesian space $\mathbb{R}^{dim\mathfrak{g}}$. This transformation is a linear map defined as $(.)^{\wedge}:\mathbb{R}^{dim\mathfrak{g}} \rightarrow \mathfrak{g}$.  
Therefore, for the vector $\boldsymbol{\zeta}=\text{vec}(\boldsymbol{\zeta}_{{R}},\boldsymbol{\zeta}_{{v}}, \boldsymbol{\zeta}_{{p}},\boldsymbol{\zeta}_{\Delta {R}},\boldsymbol{\zeta}_{\Delta {p}}) \in \mathbb{R}^{dim\mathfrak{g}}, $ this linear map has the form \cite{sola2018micro},\cite{hartley2020contact}:
\begin{equation}
    \begin{gathered}  \label{error_approx}
    \boldsymbol{\zeta}^{\wedge}=\begin{bmatrix}
    (\boldsymbol{\zeta}_{{R}})_\times & \boldsymbol{\zeta}_{{v}} & \boldsymbol{\zeta}_{{p}} & \mathbf{0}_{3,3} & \mathbf{0}_{3,1}\\
    \mathbf{0}_{1,3} & 0 & 0 & \mathbf{0}_{1,3} & \mathbf{0}_{3,1}\\
    \mathbf{0}_{1,3} & 0 & 0 & \mathbf{0}_{1,3} & \mathbf{0}_{3,1}\\
    \mathbf{0}_{3,3} & \mathbf{0}_{3,1} & \mathbf{0}_{3,1} & (\boldsymbol{\zeta}_{\Delta {R}})_\times & \boldsymbol{\zeta}_{\Delta {p}}\\
    \mathbf{0}_{1,3} & 0 & 0 & \mathbf{0}_{1,3} & 0
    \end{bmatrix} \in \mathfrak{g}.
    \end{gathered} 
\end{equation}
Now we can define the exponential map of our Lie group, 
$\boldsymbol{\eta}_t=\exp(\boldsymbol{\zeta})$.
This exponential map takes $\forall \boldsymbol{\zeta} \in \mathbb{R}^n $ to the corresponding matrix representation in $G$ as: 
\begin{equation}
    \exp(.): \mathbb{R}^{dim\mathfrak{g}}\rightarrow G, \quad  \exp(\boldsymbol{\zeta})=\exp_m (\boldsymbol{\zeta}^\wedge),
\end{equation}
where $\exp_m(.)$ is the matrix exponential.

The dynamics of the system can be written using \eqref{IMU-input-nobias}-\eqref{offset-dynamics}:
 \begin{equation}
    \begin{aligned}
    \frac{d}{dt}\mathbf{X}_t
    =&\begin{bmatrix}
\mathbf{R}_t(\tilde{\boldsymbol{\omega}}_t)_\times & \mathbf{R}_t \tilde{\mathbf{a}}_t+\mathbf{g} & \mathbf{v}_t & \mathbf{0}_{3,3} & \mathbf{0}_{3,1}\\
\mathbf{0}_{1,3} & 0 & 0 & \mathbf{0}_{1,3} & \mathbf{0}_{3,1}\\
\mathbf{0}_{1,3} & 0 & 0 & \mathbf{0}_{1,3} & \mathbf{0}_{3,1}\\
\mathbf{0}_{3,3} & \mathbf{0}_{3,1} & \mathbf{0}_{3,1} & \mathbf{0}_{3,3} & \mathbf{0}_{3,1}\\
\mathbf{0}_{1,3} & 0 & 0 & \mathbf{0}_{1,3} & 0
\end{bmatrix}\\
&-\mathbf{X}_t
\begin{bmatrix}
(\mathbf{w}_{\omega_t})_\times & \mathbf{w}_{a_t} & \mathbf{0}_{3,1} & \mathbf{0}_{3,3} & \mathbf{0}_{3,1}\\
\mathbf{0}_{1,3} & 0 & 0 & \mathbf{0}_{1,3} & \mathbf{0}_{3,1}\\
\mathbf{0}_{1,3} & 0 & 0 & \mathbf{0}_{1,3} & \mathbf{0}_{3,1}\\
\mathbf{0}_{3,3} & \mathbf{0}_{3,3} & \mathbf{0}_{3,3} & (\mathbf{w}_{\Delta R_t})_\times & \mathbf{w}_{\Delta p_t}\\
\mathbf{0}_{1,3} & 0 & 0 & \mathbf{0}_{1,3} & 0
\end{bmatrix}\\
\triangleq & {f}_{u_t}(\mathbf{X}_t)-\mathbf{X}_t \mathbf{w}_t ^\wedge, \label{Dynamics}
    \end{aligned}
\end{equation}
where $(\cdot)_t$ denotes the value of the variable $(\cdot)$ at time instant $t$.
Here the noise vector $\mathbf{w}_t$ is defined as $\mathbf{w}_t \triangleq \text{vec}(\mathbf{w}_{\omega_t},\mathbf{w}_{a_t} ,\mathbf{0}_{3,1}, \mathbf{w}_{\Delta R_t} , \mathbf{w}_{\Delta p_t}) $. 

It can be shown that the deterministic dynamics ${f}_{u_t}(.)$ meets the following group affine condition \cite{barrau2016invariant}:
\begin{equation}
{f}_{u_t}(\mathbf{X}_1 \mathbf{X}_2)
={f}_{u_t}(\mathbf{X}_1)\mathbf{X}_2 + \mathbf{X}_1 {f}_{u_t}(\mathbf{X}_2) -\mathbf{X}_1 {f}_{u_t}(\mathbf{I})\mathbf{X}_2.
\label{eqn: group affine}
\end{equation}
Therefore, according to \cite{barrau2016invariant}, the right-invariant error has deterministic autonomous dynamics (that are independent of state) as below: 
\begin{equation}
    \frac{d}{dt}\boldsymbol \eta_t= {g}_{u_t}(\boldsymbol \eta_t), \quad {g}_{u_t}( \boldsymbol \eta_t)= {f}_{u_t}(\boldsymbol \eta_t)-\boldsymbol \eta_t {f}_{u_t}(\mathbf{I}),
    \label{error-dif}
\end{equation}
and if we consider the noise in the system we will have:
\begin{equation}
\begin{gathered}
    \frac{d}{dt} \boldsymbol{\eta}_t =
    {g}_{u_t}(\boldsymbol{\eta}_t) + {Ad}_{\bar{\mathbf{X}}_t} \mathbf{w}_t^\wedge.
    \label{error_noise}
    \end{gathered}
\end{equation}
Here, for any $\mathbf X_t \in G$, the adjoint map $ Ad_{\mathbf{X}_t}:\mathfrak{g} \rightarrow \mathfrak{g} $ is the linear mapping from the local tangent space (defined at $\mathbf{X}_t$) to the global tangent space (defined at the identity element $\mathbf{I}$) in the Lie algebra, defined as
$
    Ad_{\mathbf{X}_t}(\cdot)^\wedge \triangleq \mathbf{X}_t(\cdot)^\wedge \mathbf{X}_t^{-1}
$.
Therefore, the adjoint matrix representation for $\mathbf{X}_t$ can be obtained as:
\begin{equation}
    Ad_{\mathbf{X}_t}= \begin{bmatrix}
\mathbf{R}_t & \mathbf{0}_{3,3} & \mathbf{0}_{3,3} & \mathbf{0}_{3,3} & \mathbf{0}_{3,3}\\
(\mathbf{v}_t)_\times \mathbf{R}_t & \mathbf{R}_t & \mathbf{0}_{3,3} & \mathbf{0}_{3,3} & \mathbf{0}_{3,3}\\
(\mathbf{p}_t)_\times \mathbf{R}_t & \mathbf{0}_{3,3} & \mathbf{R}_t & \mathbf{0}_{3,3} & \mathbf{0}_{3,3}\\
\mathbf{0}_{3,3} & \mathbf{0}_{3,3} & \mathbf{0}_{3,3} & \Delta \mathbf{R}_t & \mathbf{0}_{3,3}\\
\mathbf{0}_{3,3} & \mathbf{0}_{3,3} & \mathbf{0}_{3,3} & (\Delta \mathbf{p}_t)_\times \Delta \mathbf{R}_t & \Delta \mathbf{R}_t
\end{bmatrix}.
\label{adjoint_matrix}
\end{equation}
Moreover, we can obtain a log-linear error equation using the first-order approximation of the exponential map and (\ref{error_noise}).
By the definition of $\exp$, we have $\boldsymbol{\eta}_t=\exp(  \boldsymbol{\zeta}_t)\approx \mathbf{I} +  \boldsymbol{\zeta}_t^\wedge\label{linear_exp}$.
Also, by the theory of invariant filtering~\cite{barrau2016invariant}, we can obtain the Jacobian $\mathbf{A}_t$ of the deterministic portion of (\ref{error_noise}):
\begin{gather}
{g}_{u_t}(\exp( \boldsymbol{\zeta}_t))=(\mathbf{A}_t  \boldsymbol{\zeta}_t)^\wedge + \text{h.o.t.}(||  \boldsymbol{\zeta}_t||)\approx (\mathbf{A}_t  \boldsymbol{\zeta}_t)^\wedge, \label{linear_Gut} 
\end{gather}
where
$\text{h.o.t.}$ represents the higher-order terms.
Then, from \eqref{error-dif}, we obtain the log-linear error equation:
\begin{equation}
    \frac{d}{dt}  \boldsymbol{\zeta}_t=\mathbf{A}_t  \boldsymbol{\zeta}_t.
    \label{linear_dif_error}
\end{equation}
Therefore, given the initial right-invariant error $\boldsymbol{\eta}_0=\exp(\boldsymbol{\zeta}_0)$, $\boldsymbol{\eta}_t$ can be recovered using (\ref{linear_dif_error}).
This results in a linear right-invariant error propagation (prediction) in the filter, which is exact for the deterministic case.
With the process noise considered, the linear error equation in $\boldsymbol \zeta_t$ becomes $\frac{d}{dt}  \boldsymbol{\zeta}_t=\mathbf{A}_t  \boldsymbol{\zeta}_t + \textcolor{black}{{Ad}_{\bar{\mathbf{X}}_t} \mathbf{w}_t^\wedge}$.

We are now ready to derive the expression of $\mathbf{A}_t$ defined in (\ref{linear_Gut}), by substituting the first-order approximation of the right-invariant error into the definition of $g_{u_t}$ in \eqref{error_noise}:
\begin{equation}
\small
    \begin{aligned}
    & {g}_{u_t}(\exp( \boldsymbol{\zeta}_t)) \approx {g}_{u_t}(\mathbf{I} + \boldsymbol{\zeta}_t^{\wedge}) \\
    &=\begin{bmatrix}
    (\mathbf{I}_3 +(\boldsymbol{\zeta}_{R_t})_\times)(\tilde{\boldsymbol{\omega}}_t)_\times & (\mathbf{I}_3 +(\boldsymbol{\zeta}_{R_t})_\times)\tilde{\mathbf{a}}_t+\mathbf{g} & \boldsymbol{\zeta}_{v_t} & 0_{3,4}\\
    \mathbf{0}_{1,3} & 0 & 0 & \mathbf{0}_{1,4} \\
    \mathbf{0}_{1,3} & 0 & 0 & \mathbf{0}_{1,4} \\
    \mathbf{0}_{3,3} & \mathbf{0}_{3,1} & \mathbf{0}_{3,1} & \mathbf{0}_{3,4}\\
    \mathbf{0}_{1,3} & 0 & 0 & \mathbf{0}_{1,4}
    \end{bmatrix}\\
    ~~&-
\left[
   \begin{array}{c*{5}{@{\,}c}}
    \mathbf{I}_3+(\boldsymbol{\zeta}_{R_t})_\times & \boldsymbol{\zeta}_{v_t} & \boldsymbol{\zeta}_{p_t} & \mathbf{0}_{3,3} & 0\\
    \mathbf{0}_{1,3}  &  0 & 0 & \mathbf{0}_{1,3} & 0\\
    \mathbf{0}_{1,3}  &  0 & 0 & \mathbf{0}_{1,3} & 0\\
    \mathbf{0}_{3,3}  &  \mathbf{0}_{3,1} & \mathbf{0}_{3,1} & (\boldsymbol{\zeta}_{\Delta R_t})_\times & \boldsymbol{\zeta}_{\Delta p_t}\\
    \mathbf{0}_{1,3}  &  0 & 0 & \mathbf{0}_{1,3} & 0 
   \end{array} 
   \right]
   \begin{bmatrix}
    (\tilde{\boldsymbol{\omega}}_t)_\times & \tilde{\mathbf{a}}_t+\mathbf{g}  & \mathbf{0}_{3,5}\\
    \mathbf{0}_{1,3} & 0 &  \mathbf{0}_{1,5}\\
    \mathbf{0}_{1,3} & 0 &  \mathbf{0}_{1,5}\\
    \mathbf{0}_{3,3} & \mathbf{0}_{3,1} &  \mathbf{0}_{3,5}\\
    \mathbf{0}_{1,3} & 0 &  \mathbf{0}_{1,5}
    \end{bmatrix}\\
    &=  \begin{bmatrix}
    \mathbf{0}_{3,3} & (\mathbf{g})_\times \boldsymbol{\zeta}_{R_t} & \boldsymbol{\zeta}_{v_t}& \mathbf{0}_{3,4}\\
    \mathbf{0}_{1,3} & 0 & 0 & \mathbf{0}_{1,4} \\
    \mathbf{0}_{1,3} & 0 & 0 & \mathbf{0}_{1,4} \\
    \mathbf{0}_{3,3} & \mathbf{0}_{3,1} & \mathbf{0}_{3,1} & \mathbf{0}_{3,4}\\
    \mathbf{0}_{1,3} & 0 & 0 & \mathbf{0}_{1,4}
    \end{bmatrix}=\begin{bmatrix}
     \mathbf{0}_{3,1}\\ (\mathbf g)_\times \\ \boldsymbol{\zeta}_{v_t} \\ \mathbf{0}_{3,1} \\ \mathbf{0}_{3,1}
    \end{bmatrix}^\wedge,
    \end{aligned}
    \end{equation}
which yields
    \begin{equation}
    \mathbf{A}_t= \begin{bmatrix}
    \mathbf{0}_{3,3} & \mathbf{0}_{3,3}  & \mathbf{0}_{3,3}& \mathbf{0}_{3,3} & \mathbf{0}_{3,3}\\
    {(\mathbf{g})_\times} & \mathbf{0}_{3,3} & \mathbf{0}_{3,3} & \mathbf{0}_{3,3} & \mathbf{0}_{3,3} \\
    \mathbf{0}_{3,3} & \mathbf{I}_{3} & \mathbf{0}_{3,3} & \mathbf{0}_{3,3} & \mathbf{0}_{3,3}\\
    \mathbf{0}_{3,3} & \mathbf{0}_{3,3}  & \mathbf{0}_{3,3}& \mathbf{0}_{3,3} & \mathbf{0}_{3,3}\\
    \mathbf{0}_{3,3} & \mathbf{0}_{3,3}  & \mathbf{0}_{3,3}& \mathbf{0}_{3,3} & \mathbf{0}_{3,3}
    \end{bmatrix},
    \label{error_jacobian}
\end{equation}
where $\mathbf{I}_3$ is the $3\times3$ identity matrix.

Now we can write down the predication step of our InEKF, which consists of the propagation of the state estimate $\bar{\mathbf X}_t$ through the process model as well as the propagation of the covariance matrix $\mathbf P_t$ through the Riccati equation \cite{maybeck1982stochastic}: 
\begin{equation}
    \frac{d}{dt}\mathbf{\bar{X}}_t=f_{u_t}(\mathbf{\bar{X}}_t), \quad \frac{d}{dt} \mathbf{P}_t= \mathbf{A}_t \mathbf{P}_t + \mathbf{P}_t \mathbf{A}_t^T + \Bar{\mathbf{Q}}_t,
\end{equation}
where $\Bar{\mathbf Q}_t$ is the process noise covariance defined as 
$
    \Bar{\mathbf{Q}}_t
    = Ad_{{\Bar{\mathbf{X}}_t}} \text{Cov}(\mathbf{w}_t) Ad_{\Bar{\mathbf{X}}_t} \label{Q_cov}
$.

\subsection{Measurement Model and Update}
As stated in Sec.~\ref{sec:Formulation_Leg_kinematics}, we are using leg forward kinematics to get the relative position of the desired contact point (toe) in the measurement frame.
However, unlike \cite{hartley2020contact}, we cannot directly use (\ref{leg_kinmetics}) as our measurement model since we did not include the position of the contact point as part of the state.
Instead, inspired by \cite{teng2021legged}, we used the derivative of (\ref{leg_kinmetics}) as our measurement model.
Such a formulation allows the stochastic
dynamics (i.e., the noise term) in (\ref{Dynamics}) to be linear, 
so that we can achieve log-linear error dynamics \eqref{linear_dif_error} while including the offset variables.

Considering the contact point position in the world frame (i.e., $\mathbf d_t$), we have:
\begin{equation}
\begin{aligned}
    & \quad \mathbf d_t^M=(\R_t)\mathbf R_t^T (\mathbf d_t-\mathbf p_t) -\p_t ={FK}(\boldsymbol{\alpha}_t)
    \\
    \Rightarrow & \quad \mathbf d_t -\mathbf p_t =\mathbf R_t (\R_t)^T(\p_t + {FK}( \boldsymbol{\alpha}_t))
    \\
    \Rightarrow & \quad \frac{d}{dt}{(\mathbf{d}_t-\mathbf{p}_t)}
    = 
    \mathbf R_t (\R_t)^T (\mathbf w_{\Delta p_t} +  J(\boldsymbol{\alpha}_t) (\dot{\boldsymbol{\alpha}_t}+\mathbf w_{\dot{\alpha}_t}))
    \\
    & \quad +
    \left( \mathbf R_t(\boldsymbol{\omega})_\times {\mathbf R_t^T} 
    + \mathbf R_t(\R_t (\mathbf w_{\R_t})_\times)^T \right) ( {FK}(\boldsymbol{\alpha}_t)+\p_t),
\end{aligned}
\end{equation}
where $ J $ is the forward kinematic Jacobian ($ J (\boldsymbol{\alpha}) \triangleq \frac{\partial {FK} (\boldsymbol{\alpha})}{\partial \boldsymbol{\alpha}}$)
and $\mathbf{w}_{\dot{\alpha}_t}$ is the joint velocity measurement noise.

Assuming that the contact point is stationary in the world frame (i.e., $\dot{\mathbf d}_t=\mathbf 0$), knowing $\dot{\mathbf p}_t=\mathbf v_t$, and using the property of the multiplication of a skew-symmetric matrix and a vector (i.e. $(\mathbf a)_\times \mathbf b=-(\mathbf b)_\times \mathbf a$), the measurement model can be simplified as:
\begin{equation}
    \mathbf y= {h}(\mathbf X_t)+\mathbf n_t
    \label{kinematics measurement equation}
\end{equation}
where
$\mathbf y = - J(\boldsymbol{\alpha}_t)  \dot{\boldsymbol{\alpha}}_t$, ${h}(\mathbf X_t) = (\R_t)\mathbf R_t^T \mathbf v_t -  (\p_t)_\times \R_t \boldsymbol{\omega}_t - \left(  {FK}(\boldsymbol{\alpha}_t) \right)_\times \R_t \boldsymbol{\omega}_t$,
and the vector $\mathbf n_t$ contains the measurement noise terms.

Note that our measurement model is nonlinear and does not have the right-invariant observation form, meaning the innovation does not solely depends on the invariant error \cite{barrau2016invariant,hartley2020contact}.
Therefore, we used the standard EKF procedure here to formulate the innovation and update equations.

First we need to find the Jacobian of the measurement model with respect to $\boldsymbol{\zeta}_t$, denoted as $\mathbf H_t$:
\begin{equation}
    \begin{gathered}
    \mathbf{H}_t \boldsymbol{\zeta}_t + \text{h.o.t}(\boldsymbol{\zeta}_t)
    \triangleq
        {h}(\bar{\mathbf X}_t)-{h}(\mathbf X_t).  \label{measurement_error}
    \end{gathered}
\end{equation}
To express ${h}(\bar{\mathbf{X}}_t)-{h}(\mathbf{X}_t)$ in terms of $\boldsymbol{\zeta}_t$, we need to use the following first-order approximation along with (\ref{error_approx}):
\begin{equation*}
\small
    \begin{aligned}
        \boldsymbol{\eta}_t&=\bar{\mathbf{X}}_t {\mathbf{X}}_t^{-1}
        \\
      &= \left[
   \begin{array}{c*{5}{@{\,}c}}
\bar{\mathbf{R}}_t\mathbf{R}_t ^T & \bar{\mathbf{v}}_t & \bar{\mathbf{p}}_t & \mathbf{0}_{3,3} & 0\\
& \ \ -\bar{\mathbf{R}}_t\mathbf{R}_t ^T \mathbf{v}_t & \ \ -\bar{\mathbf{R}}_t\mathbf{R}_t ^T \mathbf{p}_t & &\\
\mathbf{0}_{1,3} & 0 & 0 & \mathbf{0}_{1,3} & 0\\
\mathbf{0}_{1,3} & 0 & 0 & \mathbf{0}_{1,3} & 0\\
\mathbf{0}_{3,1} & \mathbf{0}_{3,1} & \mathbf{0}_{3,1} & \overline{\R}_t \R_t ^T & \bp_t\\
 & & & &  -\overline{\R}_t \R_t^T \p_t\\
\mathbf{0}_{1,3} & 0 & 0 & \mathbf{0}_{1,3} & 0
   \end{array} 
   \right]\\
   &\approx \mathbf{I}+ \boldsymbol{\zeta}_t^{\wedge} \\ 
   {\Rightarrow}~ &
   \mathbf R_t^T \approx \bar{ \mathbf R}_t^T (\mathbf{I} +(\boldsymbol{\zeta}_{ R_t})_\times), 
   ~
   \R_t^T \approx \bR_t^T (\mathbf{I} +(\boldsymbol{\zeta}_ {\Delta R _t})_\times)\\
   & \mathbf v_t 
   \approx 
   (\mathbf{I} -(\boldsymbol{\zeta}_{ R_t})_\times)(\bar{\mathbf v}_t -\boldsymbol{\zeta}_{ v_t}),  
   \\  
   & \p_t \approx (\mathbf I- ( \boldsymbol{\zeta}_{\Delta R_t})_\times)(\overline{\p}_t -\boldsymbol{\zeta}_{\Delta p_t}).
    \end{aligned}
\end{equation*}

After ignoring the higher-order terms, we take the derivative of \eqref{measurement_error} with respect to $\boldsymbol \zeta_t$ to find $\mathbf H_t$:
\begin{equation}
\begin{gathered}
       \mathbf H_t= [\mathbf{0}_{3,3} ,
       \quad \bR_t \bar{\mathbf R_t}^T ,\quad \mathbf{0}_{3,3},\quad \mathbf h_{4} ,\quad (\bR_t \boldsymbol{\omega}_t)_\times ],\\
    \mathbf h_{4}=-(\bR_t \bar{\mathbf R}_t^T \bar{\mathbf v}_t)_\times - (\bR_t \boldsymbol{\omega}_t)_\times (\bp_t)_\times
       +(\bp_t)_\times (\bR_t \boldsymbol{\omega}_t)_\times \\
       +({FK} (\boldsymbol{\alpha}_t))_\times (\bR_t \boldsymbol{\omega}_t)_\times
\end{gathered}
\end{equation}
Similar to \cite{brossard2020ai,teng2021legged}, we can write the update equation for our InEKF as:
\begin{equation}
    \begin{gathered}
    \bar{\mathbf X}_t^+=\exp(\mathbf K_t (  \mathbf y_t-{h}(\bar{\mathbf X}_t^- ) ))\bar{\mathbf X}_t^-, \\
    \mathbf P_t^+=(\mathbf I-\mathbf K_t \mathbf H_t)\mathbf P_t^- (\mathbf I-\mathbf K_t \mathbf H_t)^T +\mathbf K_t \mathbf N_t \mathbf K_t^T,
    \end{gathered}
\end{equation}
where the Kalman gain $\mathbf K_t$ and measurement noise covariance $\mathbf N$ are defined as:
\begin{equation}
    \begin{gathered}
    \mathbf K_t 
    =\mathbf P_t \mathbf H_t ^T\mathbf S_t^{-1}, 
    \\
    \mathbf S_t=\mathbf H_t \mathbf P_t^{-}\mathbf H_t^T+\mathbf N_t,
    \quad
    \mathbf N_t=\bar{\mathbf R}_t \bR_t^T \text{Cov}(\mathbf n_t) \bR_t \bar{\mathbf R}_t^T.
    \end{gathered}
\end{equation}


\section{EXPERIMENT RESULTS}

This section introduces the experimental setup and validation results for the proposed InEKF.

\begin{figure}[h]
    \centering
    \vspace{+0.1in}
    \includegraphics[width=1\linewidth]{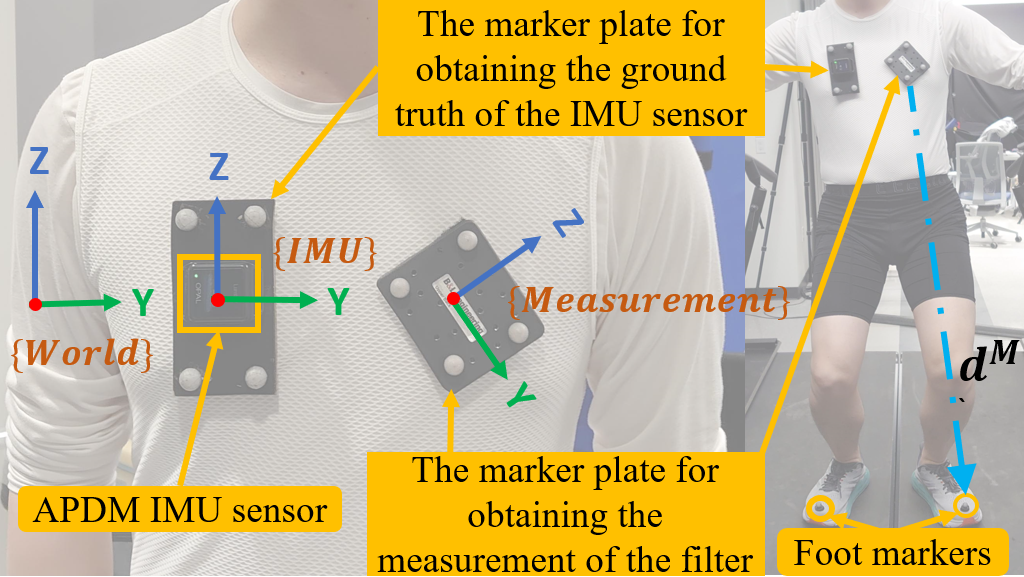}
    \caption{Experiment setup that emulates the scenario with a large offset between the IMU and measurement frames.
    The left marker plate in the left figure is rigidly attached to the IMU. 
    The right marker plate is aligned with the measurement frame. 
    The blue, dashed-line arrow indicates the 3-D contact point position in the measurement frame, which is obtained by the motion capture system.}
    \label{fig: pos vec experiment setup}
\end{figure}

Ideally, the ``perfect" placement of the IMU should be exactly aligned with the measurement frame, {because the measurement frame is one end of the stance leg's kinematic chain.}
However, in the real-world application, such a perfect placement is difficult to realize.
To demonstrate that the proposed InEKF can indeed handle large offsets between the IMU and measurement frames, we perform pilot experiments with the setup shown in Fig. \ref{fig: pos vec experiment setup}.
Note that, with this experimental setup, 
the measurement frame is now attached to the chest, instead of the pelvis center shown in Fig.~\ref{fig: parameters}.
One healthy, 24-year old male subject with height of 1.8 m and weight of 75 kg participated in the experiments. 

\subsection{Measurement model simplification}

In this pilot study, we adopt a simpler version of the proposed measurement model in \eqref{kinematics measurement equation}.
This simplification allows us to more easily collect data for validating the key aspects of the proposed filter, including the new state representation, group-affine process model, and contact-velocity based measurement model.

Instead of using the joint-angle based forward kinematics to form the measurement model, the 3-D velocity of the ground-contact point relative to the measurement frame, i.e., $\dot{\mathbf{d}}^M$, is directly used to form the measurement model.
To obtain the 3-D velocity vector for filter validation, we place a rigid plate (with four markers attached) at the chest of the subject, and choose its local frame as the measurement frame (i.e., the right plate in Fig. \ref{fig: pos vec experiment setup}).
With such simplification, the 3-D velocity vector could be obtained through a motion capture system, which is explained later with greater detail.


With the aforementioned simplification, 
the measurement model in \eqref{kinematics measurement equation} can be obtained by replacing ${FK}(\boldsymbol{\alpha}_t)$ and $J(\boldsymbol{\alpha}_t) \dot{\boldsymbol{\alpha}_t}$ with ${\mathbf{d}}^M$ and $\dot{\mathbf{d}}^M$, respectively.

\subsection{Setup for Data Collection}


\noindent \textbf{Sensors used.}
An APDM Opal IMU sensor is used to sense the subject's body (chest) movement.
The IMU, along with four markers, is fixed on a rigid plate (i.e., the left plate in Fig.~\ref{fig: pos vec experiment setup}), and the plate is attached to the subject's chest.
The position of the markers on this rigid plate is captured by eight Kestrel cameras, and is used to obtain the ground truth of the IMU pose via the Cortex software (Motion Analysis Corp.).
The other rigid marker plate in Fig.~\ref{fig: pos vec experiment setup} is used to emulate the measurement frame, and the four markers on the plate are used to get its ground truth pose.
To emulate the offset between the IMU and the measurement frames, the measurement marker plate is placed with a rotational offset of approximately 45 degrees in magnitude and a positional offset of approximately 0.12 m in magnitude.
All data were collected at 100 Hz.

\begin{table}[t]
\centering
\caption{Noise characteristics}
\begin{tabular}{ |c||c|c|  }
 \hline   
 \multirow{2}{*}{Measurement type} & Noise SD
  & Noise SD\\ &  (proposed InEKF) &  (existing InEKF) \\
 \hline
 Linear acceleration & 0.589 m/s\textsuperscript{2} &  0.5 m/s\textsuperscript{2}\\ 
 Angular velocity & 0.055~rad/s  & 0.05 rad/s\\  
 Kinematics measurement & 0.2~m/s & 0.05 m\\
 Placement offset ($\p$, $\R$) & (0.01 m, 0.01 rad) & NA \\
 Contact velocity & NA & 0.05 m/s\\
 \hline
\end{tabular}
\label{table: noise characteristics}
\end{table}

\noindent \textbf{Movement types.}
The human subject stood statically for 5 seconds and then began to continuously squat for 55 seconds.
Every squatting cycle took about 1.5 seconds. 

\noindent \textbf{Filters compared.}
To show the performance comparison between the proposed filter and the state of the art, the existing InEKF~\cite{hartley2020contact} is also evaluated. 
This existing InEKF was applied on a Cassie series bipedal robot where the IMU and measurement frames are perfectly aligned.
Yet, for human movement estimation, these two frames are often not perfectly aligned.
Here we evaluate its performance when the two frames are not aligned.
In the existing filter, the state variables are the orientation, velocity, and position of the IMU frame as well as the foot position.
It uses the contact point position with respect to the measurement frame to form the measurement model, which is in the right-invariant observation form.
Its deterministic system dynamics also possesses the group affine property. 
In contrast, while the deterministic system dynamics of our proposed filter satisfies the group affine property, our measurement model is not in an invariant observation form.

\noindent \textbf{Initial estimation errors.}
To illustrate the convergence rates of the two filters, we varied the initial estimated values of velocity $\mathbf v$ and orientation $\mathbf R$ across 50 trials.
The initial velocity and orientation estimates were sampled uniformly from -1 m/s to 1 m/s and from -30 degrees to 30 degrees, respectively.
The initial estimated values of $\p$ and $\R$ are set as zeros for all 50 trials.
This setting of initial estimates is used to validate both filters.

\noindent \textbf{Covariance settings.}
The characteristics (i.e., standard deviation (SD)) of process and measurement noises for the proposed InEKF and the state-of-the-art filter~\cite{hartley2020contact} are shown in Table \ref{table: noise characteristics}.
The noise characteristics for linear acceleration and angular velocity are obtained based on the nominal IMU specifications provided by APDM.
The covariances of linear acceleration and angular velocity for the two filters are individually tuned to ensure their respective best performances.
The kinematic measurement noise of the proposed filter is the vector $\mathbf n_t$ in (\ref{kinematics measurement equation}),
which contains the noise terms for the measurement of the position vector $\mathbf{d}^M$ and its derivative, the effect of foot slippage, and the IMU measurement.
The kinematic measurement noise of the previous filter only contains the uncertainties of the measurement of $\mathbf{d}^M$ and foot slippage, which are relatively low in our experiments.
Since the IMU placement offsets $\p$ and $\R$ are approximately constant, the noises for these two terms are chosen to be small.
Also, the contact velocity noise term only exists for the previous filter, which accounts for foot slippage.

\subsection{Results}

\noindent \textbf{Computational cost.} 
The proposed and existing filters processed the experimental data in MATLAB.
The computation time of one filter loop for both filters was approximately 1 ms.
Thus, both filters give reasonable computational loads. 

\begin{figure}[t]
    \centering
    \includegraphics[width=1\linewidth]{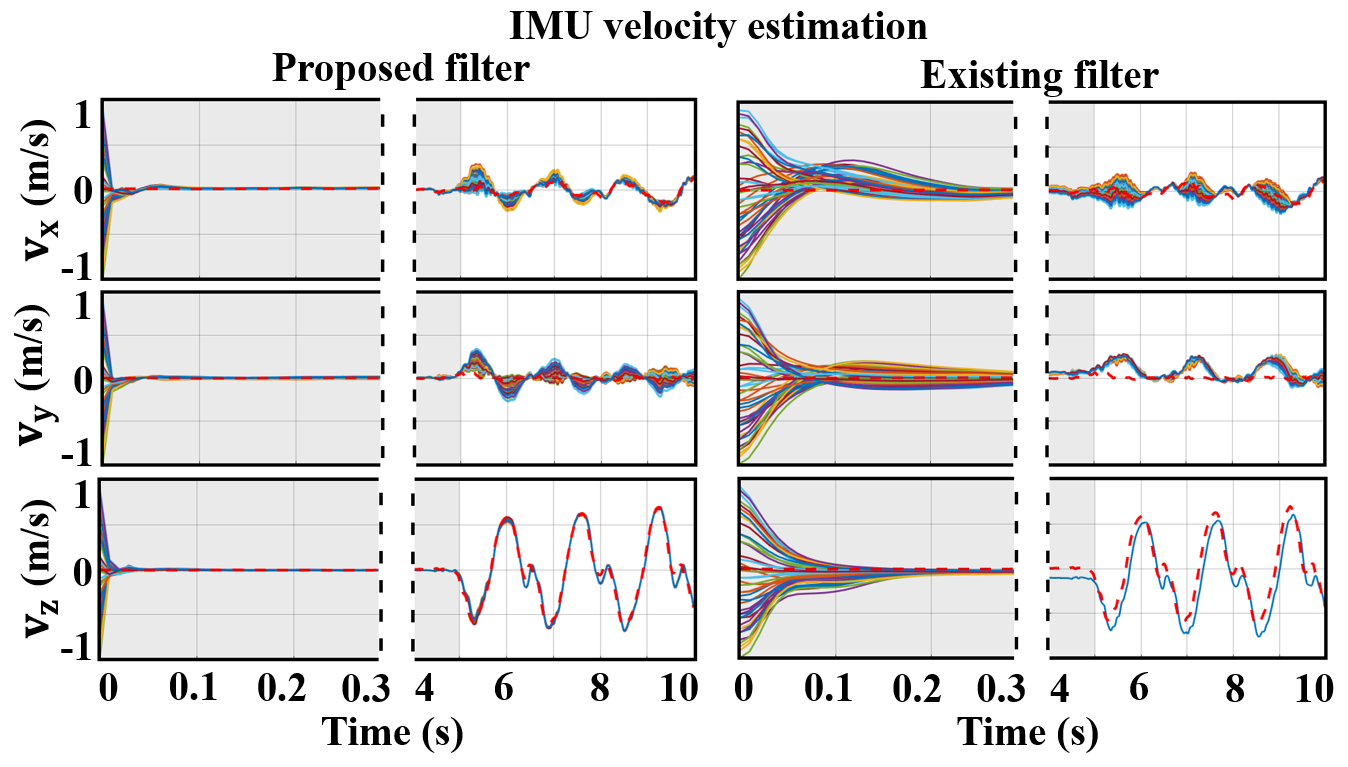}
    \caption{Estimation results of the IMU velocity ($\mathbf v \triangleq [v_x,v_y,v_z]^T$) under the proposed InEKF (left) and the existing design (right). 
    The shaded and clear backgrounds respectively indicate the periods of standing and squatting. 
    The solid and dashed lines are the estimates and the ground truth of the state variables, respectively.
    }
    \label{fig: estimation velocity w position vector}
\end{figure}

\begin{figure}[t]
    \centering
    \includegraphics[width=1\linewidth]{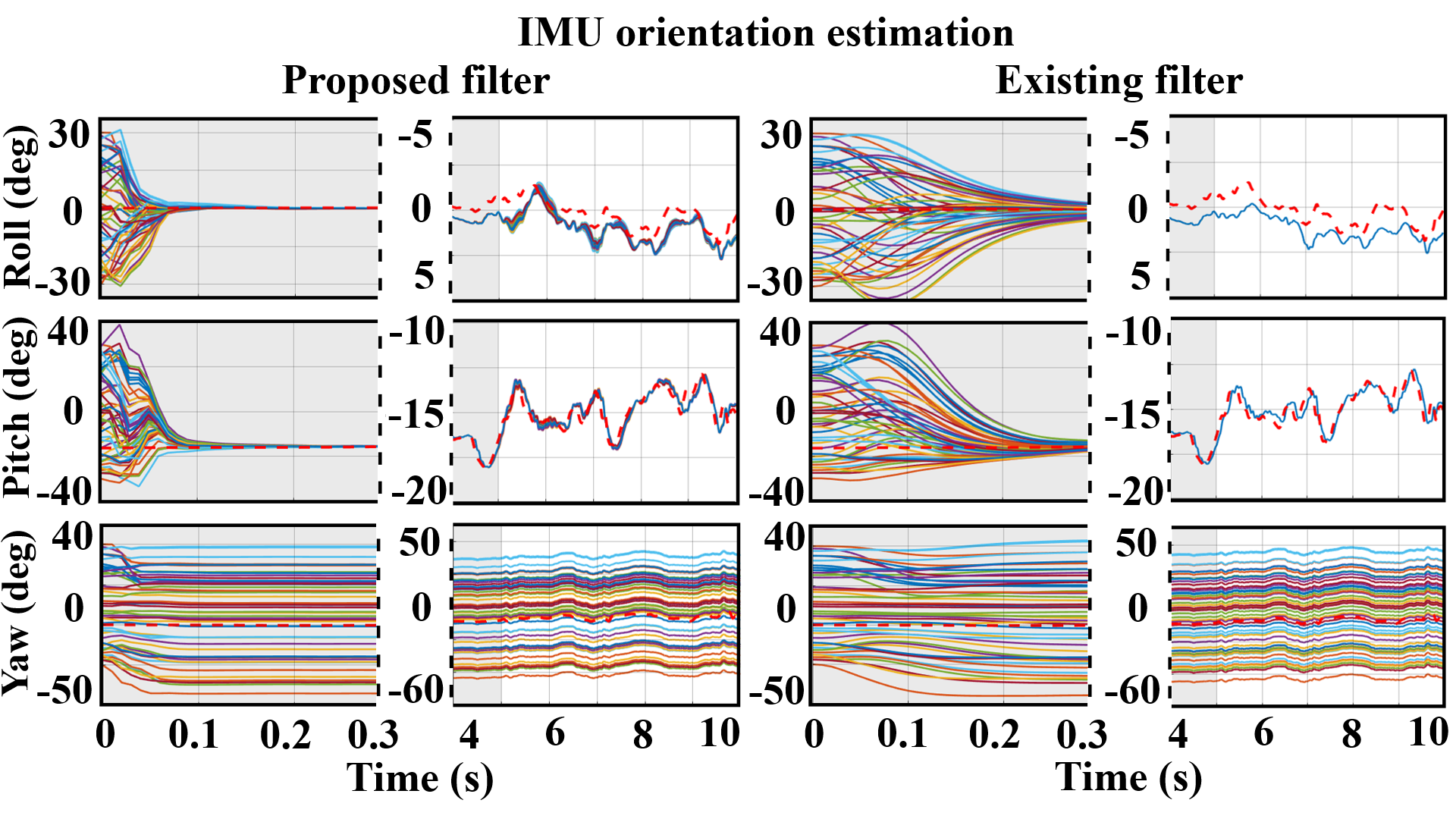}
    \caption{Estimation results of IMU orientation $\mathbf{R}$ under the proposed InEKF (left) and the existing design (right). 
    The shaded and clear backgrounds respectively indicate the periods of standing and squatting.  
    The solid and dashed lines are the estimates and the ground truth of the state variables, respectively.
    }
    \label{fig: estimation orientation w position vector}
\end{figure}

\noindent \textbf{Convergence rate.}
Figures~\ref{fig: estimation velocity w position vector} and \ref{fig: estimation orientation w position vector} display the estimation results of IMU velocity and orientation under the proposed and existing filters. 
Although both filters converge fast in the presence of large initial errors and IMU placement offset, the convergence rate of the proposed filter is 75\% faster than the existing design.
The improved convergence rate can be attributed to the higher accuracy of the proposed measurement model under large IMU placement offset, which results in more effective error correction during the update step of the proposed filter. 

\begin{figure}[t]
    \centering
    \includegraphics[width=1\linewidth]{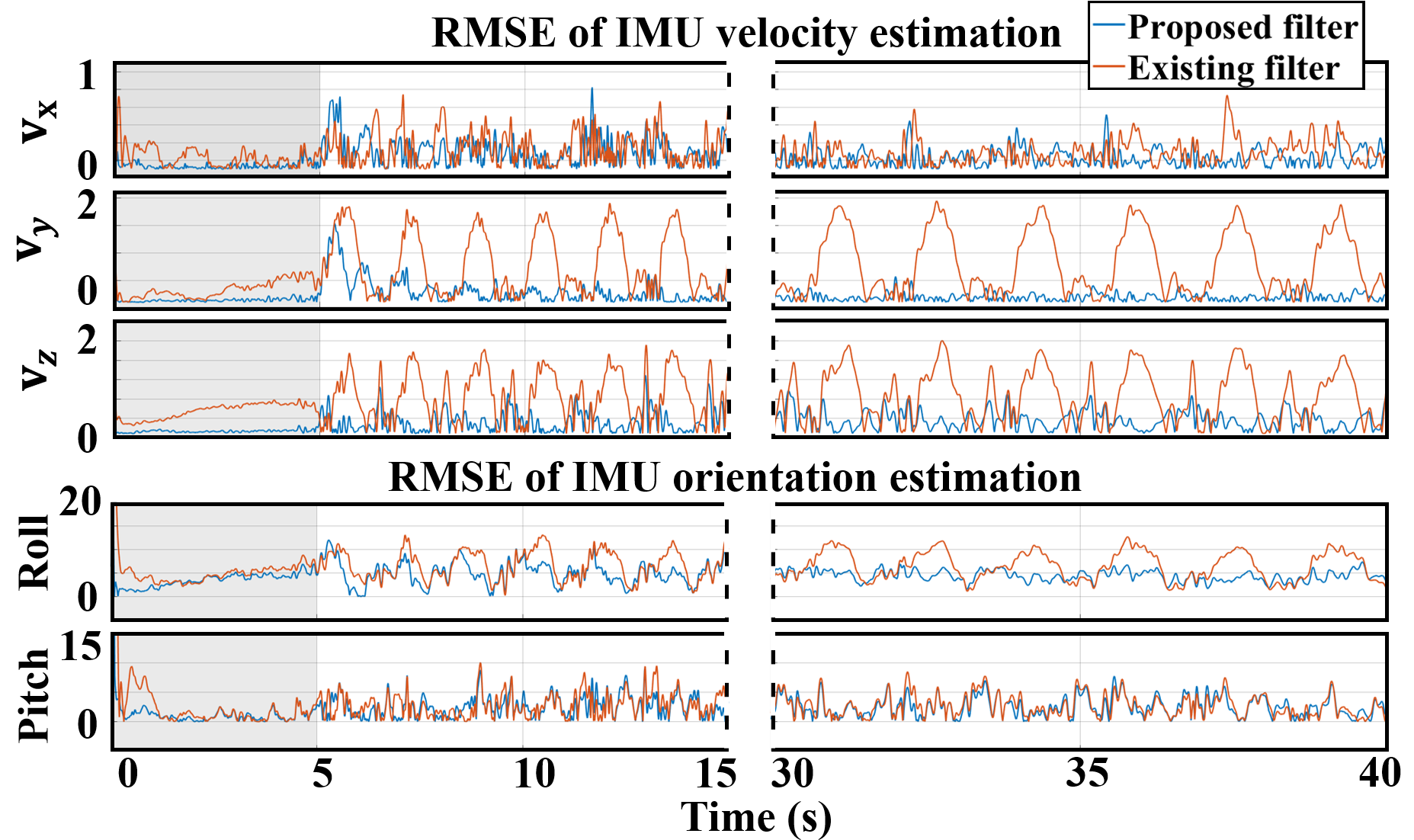}
    \caption{The RMSE of the estimated IMU velocity and orientation (only roll and pitch) for 50 trials.
    The shaded and clear backgrounds respectively indicate the periods of standing and squatting.}
    \label{fig: RMSE of velocity and orientation}
\end{figure}

\noindent \textbf{Estimation accuracy.}
The root mean square errors (RMSEs) of the estimated IMU velocity and orientation (only roll and pitch) for all 50 trials are presented in Fig.~\ref{fig: RMSE of velocity and orientation}.
While the yaw angle of the IMU frame is not observable under both filters,
the estimated roll and pitch angles converge to a small neighborhood of the ground truth under both filters.
Yet, Figs.~\ref{fig: estimation velocity w position vector} and~\ref{fig: RMSE of velocity and orientation} clearly indicate that the accuracy of the velocity estimation of the proposed filter is better than the existing filter, especially for the estimation of $v_y$ and $v_z$.

\begin{figure}[t]
    \centering
    \includegraphics[width=0.99\linewidth]{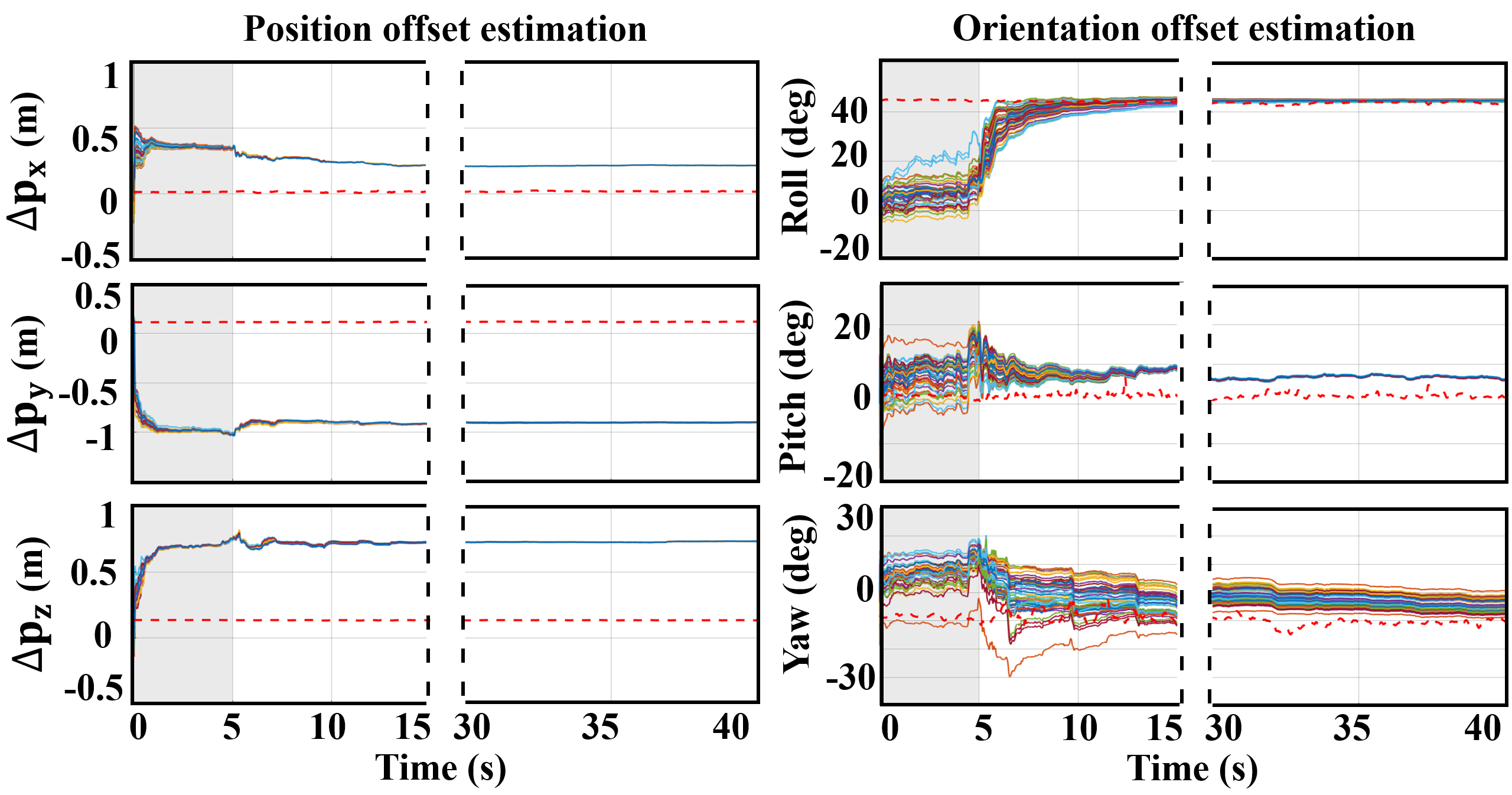}
    \caption{Estimation results of the IMU position offset $\p$ (left) and orientation offset $\R$ (right) under the proposed filter. 
    The shaded and clear backgrounds respectively indicate the periods of standing and squatting.
    The solid and dashed lines are the estimates and the ground truth of the state variables, respectively.}
    \label{fig: pos orient offset results}
\end{figure}

\noindent \textbf{Observability of IMU placement offsets.}
Figure \ref{fig: pos orient offset results} presents the estimation results of IMU placement offsets under the proposed filter.
No comparative results are shown here because the existing filter~\cite{hartley2020contact} does not explicitly treat sensor placement errors. 
When the human subject begins to squat (at $t=5$ sec), the gyroscope reading starts to give significantly larger values of $\tilde{\boldsymbol{\omega}}$, and the estimated value of $\R$ begins to converge to its ground truth, indicating $\R$ might become observable during squat.The IMU position offset ($\p \triangleq [\Delta p_x,\Delta p_y,\Delta p_z]^T$) did not converge when the subject was standing still, which might be due to the non-observability of the offset during standing.
Once the subject began to squat, the estimated value of $\Delta p_x$ started to converge to the ground truth, but those of $\Delta p_y$ and $\Delta p_z$ converged towards certain final values far from their ground truth.


\section{CONCLUSIONS}

This paper presented a right-invariant extended Kalman filter for estimating the human body movement during squatting motions. The offsets between the IMU sensor and the measurement frame (at which the kinematic measurements are provided) were explicitly considered in the filter design. The deterministic system dynamics satisfied the group affine property. Yet, the measurement model did not have the right-invariant observation form, which made the design an ``imperfect'' invariant extended Kalman filter. 

The experimental validation of human movement was performed with one human subject during repeated squat motion. The proposed filter demonstrate faster convergence and more accurate IMU velocity estimation than the state-of-the-art filter~\cite{hartley2020contact}. From the results, the rotation about the gravity vector and the IMU positions were not observable; the y-axis and z-axis components of the position offsets were not observable but detectable. The rest of the estimated states are observable and our proposed filter gives better performance than the existing InEKF. 

In our future work, instead of using a motion capture system to obtain the accurate joint angles, a suite of IMU sensors will be used to obtain joint angles of the human subject to make the results more practical. We will consider the biases in the raw data returned by IMUs in the process model. The uncertainties in joint angle measurements and the forward kinematics model will also be addressed. We will also test the filter with other types of motions and more human subjects \textcolor{black}{for a longer duration}.

\bibliographystyle{IEEEtran}
\bibliography{references.bib}

\end{document}